\documentclass[10pt,twocolumn,letterpaper]{article}

\usepackage{iccv}
\usepackage{times}
\usepackage{epsfig}
\usepackage{graphicx}
\usepackage{amsmath}
\usepackage{amssymb}
\usepackage{booktabs}  % For professional table formatting
\usepackage{graphicx}  % For adjusting column width
\usepackage{float}
\usepackage{caption}
\usepackage{booktabs}
\usepackage{multirow}
\usepackage{makecell}
\usepackage{xspace}
\newcommand{\lras}{\textbf{\texttt{LRAS}}\xspace}

\newcommand{\lrasflow}{$\textbf{\texttt{LRAS}}_{\textbf{\texttt{FLOW}}}$\xspace}

\newcommand{\lrasrgb}{$\textbf{\texttt{LRAS}}_{\textbf{\texttt{RGB}}}$\xspace}

\newcommand{\flow}{\textbf{\texttt{FLOW}}}
\newcommand{\rgb}{\textbf{\texttt{RGB}}}

\newcommand{\bvd}{$\textbf{\texttt{BVD}}$\xspace}

\definecolor{iccvblue}{rgb}{0.21,0.49,0.74}
\usepackage[pagebackref,breaklinks,colorlinks,allcolors=iccvblue]{hyperref}
\def\paperID{15315}
\def\confName{ICCV}
\def\confYear{2025}
\title{3D Scene Understanding Through \underline{L}ocal \underline{R}andom \underline{A}ccess \underline{S}equence Modeling}

\author{
Wanhee Lee$^{1}$\footnotemark[1] \quad
Klemen Kotar$^{1}$\footnotemark[1] \quad
Rahul Mysore Venkatesh$^{1}$\footnotemark[1] \quad
Jared Watrous$^{1}$\footnotemark[1] \quad
Honglin Chen$^{2}$\footnotemark[1] \\
Khai Loong Aw$^{1}$ \quad 
Daniel L. K. Yamins$^{1}$\\
$^{1}$Stanford University,
$^{2}$OpenAI
}

\begin{document}
% \begin{document}
\twocolumn[{
    \maketitle
    \vspace{-1cm}
    \begin{center}
        \centering
        \captionsetup{type=figure}
        \captionsetup{labelfont=bf}
        \includegraphics[width=0.98\textwidth]{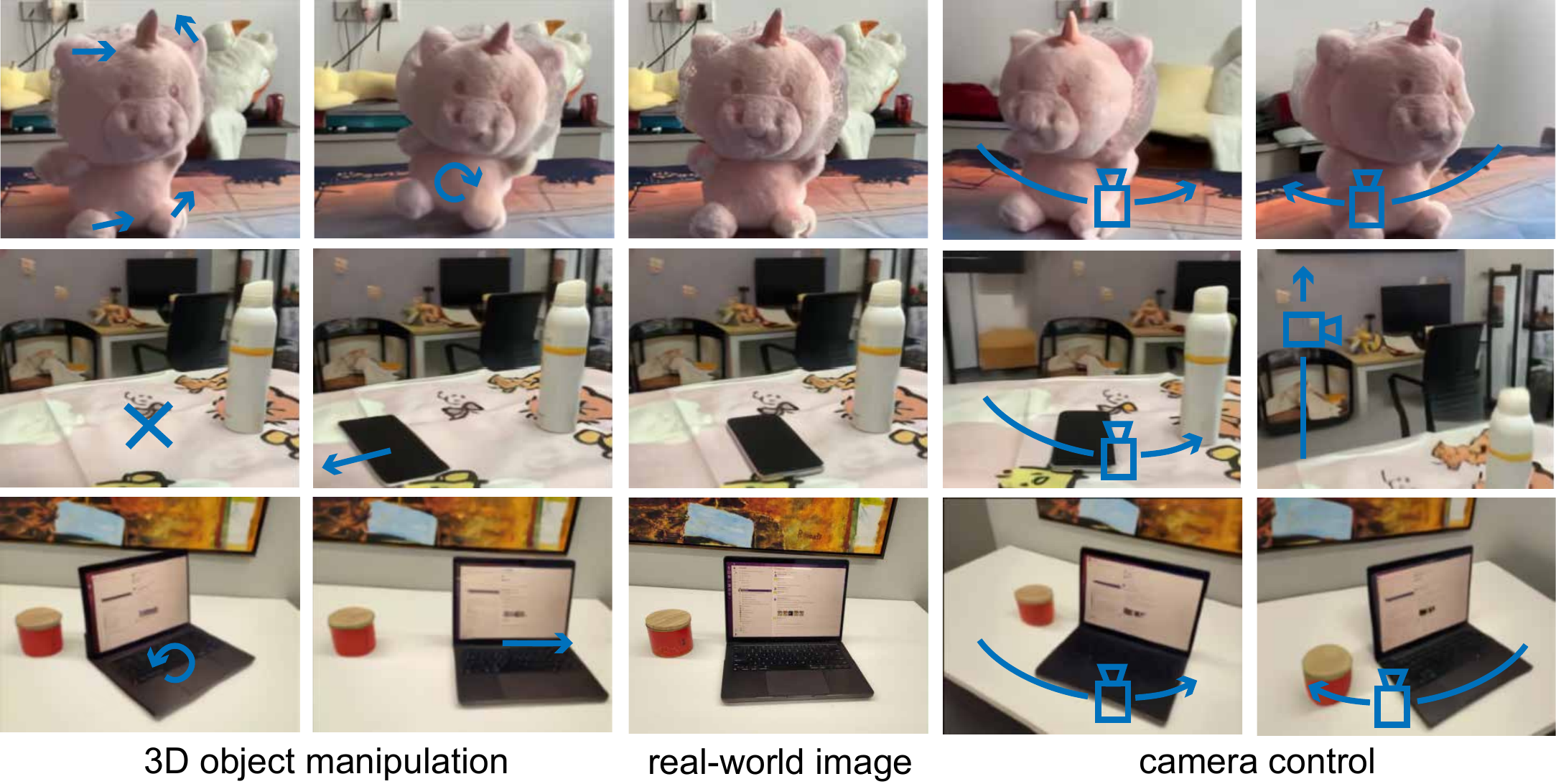}
        \caption{\textbf{Our model (\lras) enables sophisticated object manipulation (left) and camera control (right) on real-world images (middle).} The generation exhibits understanding of 3D scene structure and properties of the visual scenes, such as lighting, shadows, continuity, occlusions, and amodal completion.}
        \label{fig:teaser}
    \end{center}
    % \vspace{0.3cm}
    \vspace{-5pt}
    }]
\maketitle

% this includes maketitle

\footnotetext[1]{Equal contribution to this work.}
\footnotetext[2]{Project website at: \url{https://neuroailab.github.io/projects/lras_3d/}}

%%%%%%%%% ABSTRACT
\begin{abstract}
    % \vspace{-5pt}
3D scene understanding from single images is a pivotal problem in computer vision with numerous downstream applications in graphics, augmented reality, and robotics. While diffusion-based modeling approaches have shown promise, they often struggle to maintain object and scene consistency, especially in complex real-world scenarios. To address these limitations, we propose an autoregressive generative approach called Local Random Access Sequence (\lras) modeling, which uses local patch quantization and randomly ordered sequence generation. By utilizing optical flow as an intermediate representation for 3D scene editing, our experiments demonstrate that \lras achieves state-of-the-art novel view synthesis and 3D object manipulation capabilities. Furthermore, we show that our framework naturally extends to self-supervised depth estimation through a simple modification of the sequence design. By achieving strong performance on multiple 3D scene understanding tasks, \lras provides a unified and effective framework for building the next generation of 3D vision models.\footnotemark[2]

% empirically founding of better than diffusion + suggesting unified framework for 3D vision ->

\end{abstract}

%%%%%%%%% BODY TEXT
\vspace{-5.0mm}
\section{Introduction}

\vspace{-5pt}

Understanding 3D scenes from a single image remains a fundamental yet unsolved challenge in computer vision, and a necessary prerequisite for many robotics tasks. In this work, we study 3D scene understanding in the context of three tasks: a) Novel view synthesis -- understanding how the scene changes when the camera moves, b) 3D object manipulation -- which tests the model's ability to predict object appearance changes under rigid transformations, and c) Depth estimation -- asking how well the model perceives the 2.5D structure of visible regions.

The most dominant approaches to solving novel view synthesis and object manipulation tasks have been fine-tuning large diffusion models pre-trained on text to image or video generation \cite{wang2024motionctrl, yu2024viewcrafter, pandey2024diffusion}. Although these methods demonstrate strong capabilities and produce photorealistic images, they have certain key limitations. We benchmark the performance of these methods on in-the-wild data and find that they often fail to preserve object identity, shift global lighting, and provide imprecise control over camera and object motion. Additionally, several of these models~\cite{pandey2024diffusion, yu2024viewcrafter} rely on separate off-the-shelf supervised depth estimation models as part of their editing pipeline, sidestepping the question of how depth estimation can emerge in these large pre-trained models.

% raising concerns about whether they truly possess 3D perception alongside their generative capabilities—both of which are essential for a foundational 3D model.

% Additionally, several of these models~\cite{pandey2024diffusion, yu2024viewcrafter} rely on separate supervised depth estimation models as part of their editing pipeline, raising concerns about their inherent 3D perception capabilities beyond just generation.

% Additionally, we observe all the models need depth estimator / but do not talk about how to get that.

% A common source of these issues is the text conditioning signals used to pre-train these models, which have been found to poorly encode object identity and scene conditions \cite{Kotar2023AreTT}, resulting in them being entangled in the embedding space. While works such as \cite{Kocsis2024LightItIM, Avrahami2023TheCO} address some of these problems, our evaluations show that such issues still persist in diffusion-based scene editing models. 

%why diffusion models fail -- object identity changes becauase of reliance on clip score (). global illumination changes -- scene embedding stays same. %we train our models from scratch without reliance on clip embeddings and no text conditioning (in the default) and require complex mechanisms for preserving identity of the input image, etc (cite)  

%and vision. 

%~\cite{sun2024autoregressive, yu2023language}

%in this work, we explore an alternative approach apply autoregressive sequence modeling to this problem. 

An alternative approach is to use LLM-inspired~\cite{brown2020language} autoregressive next-token prediction for generative image modeling. Recently, such models~\cite{sun2024autoregressive, yu2023language} have emerged as a strong alternative to diffusion, outperforming diffusion baselines on image generation tasks. Unlike diffusion models, which build images from the ``top down", by turning noise into rough outlines of shapes and then filling in the detailed textures, autoregressive models build up an image from the ``bottom up", predicting the image patch by patch. However, in practice, most autoregressive models predict sequences of globally encoded tokens. In addition, these models predict sequences in raster order, allowing ``top left" tokens to have greater causal control of the predicted image, which leads to inferior generation~\cite{li2025autoregressive}. 
%-- which recent work has found -- leads to poor generation quality~\cite{li2025autoregressive}.

% Autoregressive models build up an image from the "bottom up", predicting the image patch by patch. However, in practice most autoregressive models predict sequences of globally encoded tokens each of which contain global image information. 

% In this work, we introduce a few key innovations that improve autoregressive image models and demonstrate their application to 3D scene understanding tasks. 

Our model, \lras (\textbf{\underline{L}}ocal \textbf{\underline{R}}andom \textbf{\underline{A}}ccess \textbf{\underline{S}}equence Model) addresses these shortcomings in  autoregressive image modeling and gets its name from the two key innovations: a) \textbf{\underline{L}}ocal patch representations and b) \textbf{\underline{R}}andom order decoding. In our model we predict a sequence of local patch representations which is more in line with the standard autoregresive next token prediction paradigm in LLMs. Next, we introduce architectural innovations that equip the model to decode the image in spatially random order by predicting a sequence of (pointer, contents) representations -- where the pointer indicates the spatial location at which the contents should be placed.

We take this model, and apply it to 3D understanding tasks by using optical flow intermediates. First, using \lrasrgb we learn to predict RGB images conditoned on an input frame and an optical flow map. We demonstrate that this model possesses emergent 3D scene editing capabilites such as NVS and 3D object manipulation. Further, we find that the \lras framework is flexible and can also be used as a camera conditioned flow predictor (\lrasflow) which we use to extract 2.5D depth, addressing the challenge of emergent self-supervised depth extraction from large pre-trained models. To train our model, we crawl a dataset of 7k hours of high-quality, diverse internet videos called Big Video Dataset (\bvd). We demonstrate that \bvd can be used to train powerful generative models. 

% We found this and that amazing results.
We provide empirical evidence showing the effectiveness of our approach across multiple 3D vision tasks. For novel view synthesis, our method achieves state-of-the-art performance on both object-centric and scene-level datasets. Further, to assess object manipulation capabilities, we introduce \textbf{\texttt{3DEditBench}}, a new real-world object editing benchmark. Our evaluation demonstrates that our model outperforms competing object manipulation methods on real-world data. Notably, \lras exhibits a significant advantage over diffusion models in preserving scene structure, object identity, and global illumination during 3D editing tasks. We also find that our model achieves state-of-the-art self-supervised depth estimation results on standard benchmarks on both static and dynamic objects, which has previously been hard to achieve with geometric consistency methods~\cite{zhou2017unsupervised,sun2023sc}. In this way, \lras emerges as a foundational model of 3D vision with a wide range of capabilities.

\begin{figure*}[t]
    \centering
    \includegraphics[width=0.98\linewidth]{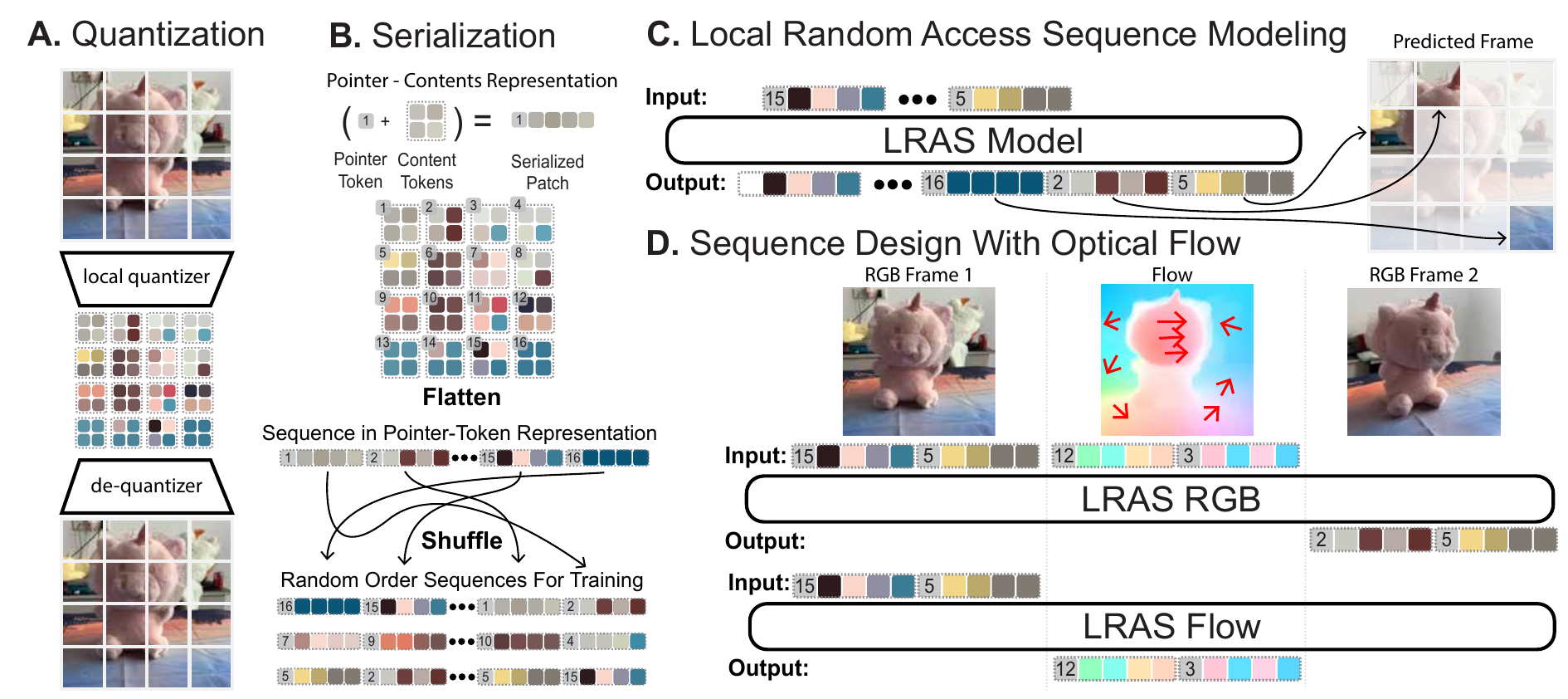}
    \captionsetup{labelfont=bf}
    \caption{\textbf{\lras Architecture.} 
    \textbf{A. Quantization:} We train a small, patch local, convolutional autoencoder with a 16 bit LFQ codebook. 
    \textbf{B. Serialization:} We serialize the codes into sequences using the pointer-content representation, which allows us to arbitrarily order the patches during training and generation.
    \textbf{C. Local Random Access Sequence Modeling}: We train an LLM-like autoregressive transformer to predict the contents of the next patch, shuffled in random order.
    \textbf{D. Sequence Design With Optical Flow}: We design sequences of tokens that contain optical flow intermediates, to provide robust control over the generation. We train two models: $\lras_\rgb$, which is conditioned on a source RGB image and an optical flow describing the desired transformation to predict the next frame, and $\lras_\flow$, which is conditioned on a source RGB image to predict a plausible optical flow field.
    }
    \label{fig:architecture}
    \vspace{-0.4cm}
\end{figure*}
% \vspace{-10pt}

% \input{figures/3Dediting_v1.tex}

\section{Related Works}

\textbf{Novel View Synthesis} (NVS) has been widely studied as a fundamental task in 3D vision. Regression-based approaches \cite{yu2021pixelnerf,charatan2024pixelsplat,kulhanek2022viewformer,sajjadi2022scene} perform well for view interpolation but struggle with single-image-to-3D synthesis, producing blurry results due to uncertainty in occluded regions. This limitation has driven a shift toward generative models, particularly diffusion-based methods, which enable high-quality and diverse NVS. Zero-1-to-3 \cite{liu2023zero}, trained on large-scale synthetic datasets \cite{deitke2023objaverse,chang2015shapenet}, predicts novel views from a single image using implicit camera modeling. 
ZeroNVS \cite{sargent2023zeronvs} integrates Zero-1-to-3’s approach with a score distillation sampling framework \cite{poole2022dreamfusion}, and extends the application to real-world scenes. 
Other approaches, such as MotionCtrl \cite{wang2024motionctrl}, inject camera embeddings to guide video diffusion without explicit 3D representations. Recently, ViewCrafter \cite{yu2024viewcrafter} utilized point-cloud rendering using DUSt3R \cite{wang2024dust3r} for improved performance with better camera motion control. In this work, we explore autoregressive sequence modeling for the NVS problem as an alternative to diffusion-based approaches to overcome the limitations of previous works.

% DreamFusion uses Score Distillation Sampling to distills knowledge from text-to-image diffusion models while enforcing 3D consistency. 

\textbf{3D Object Manipulation}
While NVS focuses on generating novel views of the input scene, object manipulation refers to the task of transforming objects in the scene while keeping the camera fixed. Drag-based image editing methods~\cite{wang2024motionctrl, draganything, shi2024lightningdrag, dragnuwa} aim to solve this problem by parameterizing object transforms as 2D motion vectors which are then used as conditioning to fine-tune stable diffusion (SD)~\cite{rombach2022high}. These methods can be naturally extended to more complex 3D transforms by incorporating depth information into the drag vectors~\cite{objctrl25D}. Another class of models~\cite{pandey2024diffusion, koo2025videohandles}, performs 3D object manipulations by editing input depth maps according to the desired object transform and utilizing a depth-conditioned diffusion model to generate the edited image. However, these methods heavily rely on inverting the input image into the SD latent space, which often fails on real-world images~\cite{mokady2023null}. In contrast, our model \lras is an autoregressive sequence model trained from scratch on internet videos, which does not rely on SD and is free from inversion processes. We find that it generalizes better to real-world images and makes more accurate edits compared to prior approaches. 

% \vspace{-1mm}

\textbf{Concurrent Work} Recently, several works that have shown that motion-conditioned diffusion models can be used to perform sophisticated image manipulations.~\cite{geng2024motion, koroglu2024onlyflow, jin2025flovd} trains a spatio-temporal trajectory-conditioned control net on top of a large video diffusion model~\cite{bar2024lumiere}. The model demonstrates emergent capabilities such as object and camera control and drag-based image editing and motion transfer. Another set of recent work~\cite{gu2025diffusion, zhang2024world, feng2024i2vcontrol} uses 3D point trajectories providing more powerful control over image generation. \lras is the first model that explores the idea of using motion conditioning to train autoregressive image generation models. We find that our model demonstrates strong performance on NVS and object manipulation tasks compared to its diffusion model counterparts.

\vspace{-1.5mm}

\section{Method}

\subsection{LRAS Architecture}

The Local Random Access Sequence Model is an autoregressive transformer with two key properties - locality and random access. \textbf{Locality} is achieved by utilizing an image quantizer which produces a grid of codes that only contain information from their corresponding patch of the input image (as illustrated in Figure \ref{fig:architecture}A). This way each token is independent of all others in the sequence - a property which is naturally present in most text tokenization schemes employed by LLMs, but is missing from modern image quantizers such as VQ-GAN \cite{Esser2020TamingTF} or the COSMOS tokenizer \cite{agarwal2025cosmos}. We hypothesize that, as with language, this gives the sequence model stronger downstream compositional abilities, as it learns to model objects as groups of individual tokens without any global dependencies on other objects or the scene.
\textbf{Random Access} is achieved by the addition of pointer tokens to the sequence (as illustrated in Figure \ref{fig:architecture}B). Since autoregressive transformers operate over 1D sequences, they have to process information in a serial order. The addition of pointers allows them to arbitrarily jump around the sequence, filling in parts of the data structure in any order. Additionally, the pointers themselves could be predicted, allowing the model to drive the generation order; we leave this exploration for future work.

\subsection{Local Patch Quantization}
Our tokenizer is a fully convolutional autoencoder with 40M parameters, with a 16 bit LPQ Bottleneck \cite{Yu2023LanguageMB} for discretizations, giving us a vocabulary size of 65,536. Our encoder consists of three ResNet \cite{He2015DeepRL} style blocks. The first layer has kernel size 4 and stride of 4, reducing the image to a 64x64 grid feature map, while subsequent encoder layers have kernel sizes and strides of 1. This design enforces that no information is shared between adjacent input patches. After the quantization layer, we apply a convolutional decoder with 6 ResNet style blocks and a kernel size 3 and stride of 1. This allows for some local information sharing between adjacent patches to make the reconstructed image coherent. The model is supervised using only L2 regression loss, and is trained on frames from the Kinetics400 \cite{kay2017kinetics} dataset. We train a second encoder, identical in architecture, to quantize optical flow fields with a 32,768-token vocabulary, using RAFT \cite{teed2020raft} flow from Kinetics400.

\subsection{Random Access Through Pointer-Contents Representation}

After quantization, an image needs to be serialized into a 1D sequence. Unlike text, which naturally follows this format, images and videos require a certain chosen order. Traditional autoregressive models use a fixed scanning order, but as shown in \cite{li2025autoregressive}, this is suboptimal. Instead, we allow the model to generate in arbitrary order. While \cite{pannatier2024sigmagpts} and \cite{Li2025FractalGM} concurrently achieve this by passing two positional embeddings to each token - one for the current token and one for the next token to generate, we take an alternative approach - the \textbf{Pointer Token}. These special tokens guide the model across the entire sequence by allowing it to ``jump'' to a new location during encoding or generation. Each pointer token is followed by the \textbf{Content Tokens}, which contain the actual RGB or Flow information at that location. During training, this allows us to randomly shuffle the order in which images are decoded, and train on only subsets of the image patches - since the image generation problem gets easier the more patches we reveal, and thus the supervision on the latter tokens is less useful. At test time, this allows us to control the order in which we predict the image, as well as only predict parts of the image or perform some of the prediction in parallel.

\subsection{Optical Flow Conditioning}
While the \lras formulation is fairly general, in this work we focus on applying it to 3D scene understanding, by utilizing optical flow intermediates. As shown concurrently in \cite{pandey2024diffusion, Gu2025DiffusionAS, Shi2024MotionI2VCA}, this formulation allows us to express any physical scene edit in the space of flow fields, yielding precisely conditioned RGB generations with diverse hallucinations. We utilize optical flow as conditioning, and uniquely also a prediction target. We illustrate how this approach naturally fits with autoregressive models, and obtain state-of-the-art results on a number of challenging 3D scene editing tasks. We introduce two models (as shown in figure \ref{fig:architecture}D):

$\lras_\rgb$ is a 7B parameter model, which takes as input an RGB frame and a dense flow field (both quantized by a local patch quantizer) and predicts the next RGB frame. $\lras_\flow$, is a 1B parameter model with the same architecture, which utilizes the flow tokens as its target, and is conditioned only on the first frame (and when available in the data, a camera pose change signal). Next, we will describe how we use \lrasrgb for 3D scene editing and \lrasflow for depth extraction. 

\subsection{Dataset and Training}
\lras was pre-trained on a large dataset containing internet videos, called \bvd (big video dataset), along with 3D vision datasets including the train splits of ScanNet++ \cite{yeshwanth2023scannet++}, CO3D \cite{reizenstein2021common}, RealEstate10K \cite{zhou2018stereo}, MVImgNet \cite{yu2023mvimgnet}, DL3DV \cite{ling2024dl3dv}, and EgoExo4D \cite{grauman2024ego} dataset. We used RAFT \cite{teed2020raft} to compute the optical flow from the videos for the training. Further information on the dataset can be found in the supplementary materials.

Our models were trained in an autoregressive fashion with cross-entropy loss applied on next token prediction. For $\lras_\rgb$, only the next frame RGB token targets are supervised. For $\lras_\flow$, only the flow tokens are supervised. Each model is optimized for 500,000 steps with a batch size of 512.

\subsection{Model Inference}
% \subsection{3D Scene Editing}
\begin{figure}[t]
    \centering
    \includegraphics[width=0.98\columnwidth]{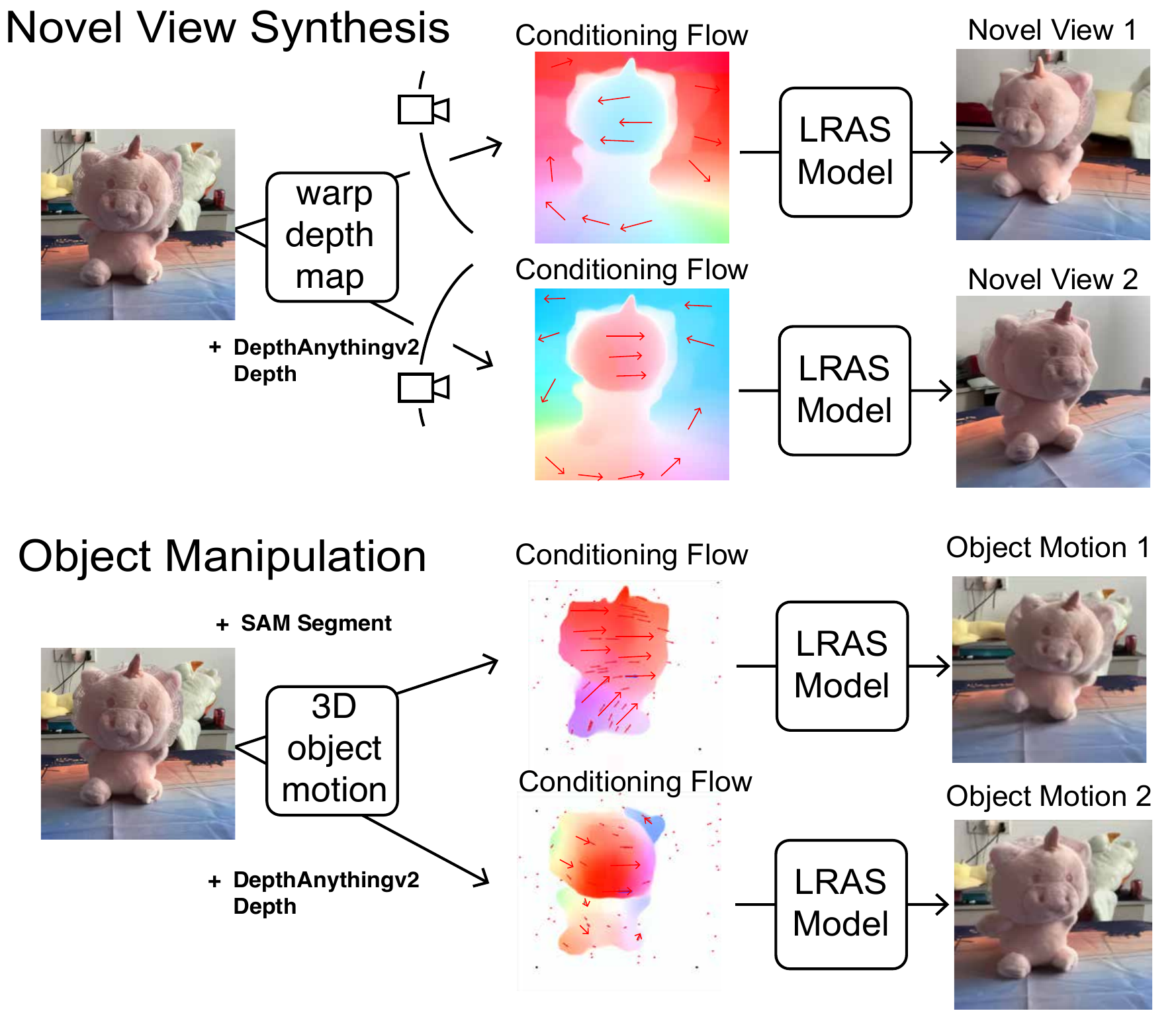}
    \captionsetup{labelfont=bf}
    \vspace{-0.2cm}
    \caption{ 
    \textbf{3D Scene Editing Through Flow Field Manipulation:} We perform 3D scene edits by constructing optical flow fields corresponding to the desired transformations - either camera or object motion in 3D.
    }
    \label{fig:3Dediting}
    % \vspace{-40pt}
    \vspace{-0.4cm}
\end{figure}
% \vspace{-10pt}
\textbf{Novel View Synthesis} can be performed using \lrasrgb by conditioning the model on 2D optical flow fields that represent how the pixels move given a desired camera pose change. To generate these flow fields, we use the following steps: a) unproject the depthmap of the input image to obtain a 3D point cloud, b) apply a rigid transformation to the point cloud as per the given camera transformation, c) re-project the transformed point cloud and compute the displacement relative to the pixels of the first frame to compute the 2D flow (See Figure~\ref{fig:3Dediting}). \lrasrgb generates the edited image given the computed flow map and the input image. As we will describe in the next section, \lrasflow provides a natural method of extracting depth maps in a self-supervised manner. However, in practice we find that marginally better performance can be achieved using off-the-shelf supervised metric depth estimators such as Depth Anything V2 \cite{yang2024depth}.

% \textbf{Novel View Synthesis} To perform NVS task using \lras, we first ran off-the-shelf depth model, DepthAnythingV2 \cite{yang2024depth}, to compute the depth. Then, we compute the 3D flow using the depth and camera pose change, and project it back to the 2D image plane to get 2D flow. Although the flow is 2D, the global flow field can give enough cue for the model to understand that it is induced by the camera motion. To resolve the camera scale ambiguity issue, we obtained a single scale value per each scene to adjust the camera translation scale, by matching the optical flow computed from the depth and camera pose and RAFT flow of the video. Then, given the first image tokens and the flow tokens, our autoregressive model performs novel view synthesis.

\textbf{3D Object Manipulation} can be performed by creating a flow field where the flow on the surface of the object characterizes the 3D transformation to be performed, with the flow of the background set to 0 -- conditioning the predictor to move the object, but keep the background fixed. We follow a similar procedure described above to produce flow fields for rigid object transformations and use the SegmentAnything \cite{kirillov2023segment} model to suppress the flow of the background regions (See Figure~\ref{fig:3Dediting}).

% \subsection{Depth Extraction and End-to-End NVS}
\textbf{Depth Extraction and End-to-End NVS}
Camera conditioned \lrasflow provides a natural method for extracting depth maps without additional finetuning. We provide in-plane camera motion as input to \lrasflow and predict the optical flow induced by camera motion. We compute the magnitude of the optical flow to compute the disparity which, when inverted, yields 2.5D depth maps. In practice, we find that a simple upward camera translation is sufficient to generate high-quality depth maps. Additionally, performance can be improved by statistical aggregation over disparity maps generated with different seeds for the same image. These depth maps can be used in conjunction with \lrasrgb for end-to-end NVS without relying on off-the-shelf depth estimators.

\section{Results}

\subsection{Novel View Synthesis}\label{subsec:nvs}

% \twocolumn[{
%     \maketitle
%     \begin{center}
%         \centering
%         \hspace*{-0.5cm}\includegraphics[width=\textwidth]{figures/nvs.png}
%         \caption{Qualitative comparison on novel view synthesis on DL3DV-11K and WildRGB-D dataset.}
%         \label{fig:nvs}
%     \end{center}
%     \vspace{0.3cm}
% }]

\begin{figure*}
    \centering
    \includegraphics[width=0.97\textwidth]{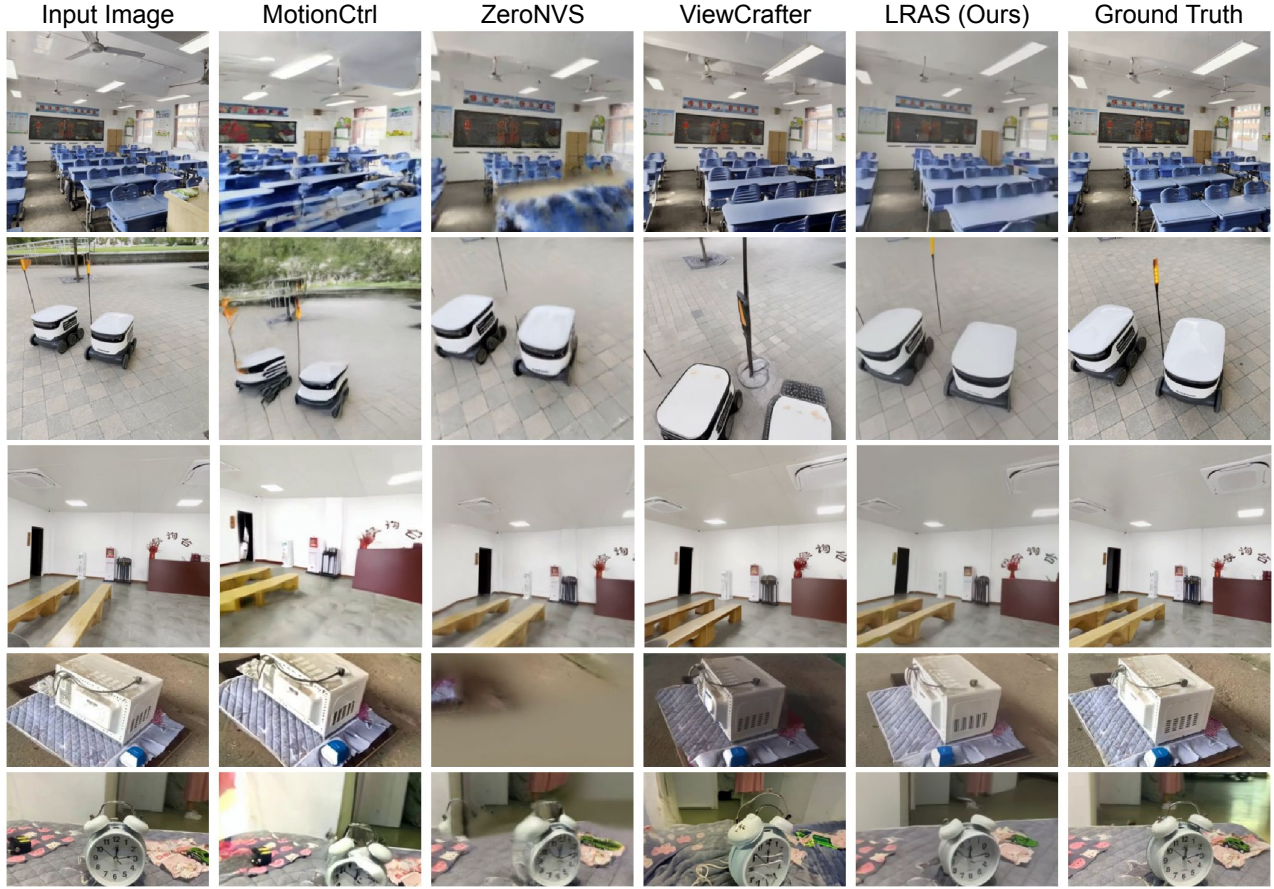}
    \captionsetup{labelfont=bf}
    \caption{\textbf{Novel view synthesis from a single image.} The results show that our model performs controllable novel view synthesis with various camera motions in a diverse scenes. Compared to other models, the reconstructed images do not show abrupt change in object and scene identity. See supplementary for more results.}
    \label{fig:nvs}
    \vspace{-0.4cm}

\end{figure*}
% \vspace{-10pt}

% Qualitative comparison on novel view synthesis from a single image on DL3DV and WildRGB-D dataset. 

% \begin{table}[t]
%     \centering
%     \renewcommand{\arraystretch}{1.2} % Adjust row height for readability
%     \setlength{\tabcolsep}{6pt} % Adjust column spacing
%     \begin{tabular}{lccc}
%         \toprule
%         \textbf{Dataset} \\
%         \textbf{Model} & \textbf{PSNR} $\uparrow$ & \textbf{SSIM} $\uparrow$ & \textbf{LPIPS} $\downarrow$ \\
%         \midrule
%         \textbf{WildRGB-D} \\
%         MotionCtrl & 13.117 & 0.321 & 0.355 \\
%         ZeroNVS  & 16.418  & 0.470  & 0.276 \\
%         ViewCrafter  & 14.138  & 0.383  & 0.280 \\
%         \textbf{LRAS(Ours)} & \textbf{17.895}  & \textbf{0.541} & \textbf{0.231} \\
%         \midrule
%         \textbf{DL3DV} \\
%         MotionCtrl & 12.629 & 0.261 & 0.462 \\
%         ZeroNVS  & 15.950  & 0.417  & 0.323 \\
%         ViewCrafter  & 16.738  & 0.437  & \textbf{0.247} \\
%         \textbf{LRAS(Ours)} & \textbf{18.279} & \textbf{0.531} & 0.321 \\
%         \bottomrule
        
%     \end{tabular}
%     \captionsetup{labelfont=bf}
%     \caption{\textbf{Comparison of metrics for novel view synthesis.}}
%     \label{tab:nvs}
% \end{table}

\begin{table}[t]
    \centering
    \hspace*{-0.3cm} % Adjust the value as needed
    \renewcommand{\arraystretch}{1.2} % Adjust row height for readability
    \setlength{\tabcolsep}{5pt} % Adjust column spacing
    % \captionsetup{labelfont=bf}
    \begin{tabular}{clccc}
        \toprule
        \textbf{Dataset} & \textbf{Model} & \textbf{PSNR} $\uparrow$ & \textbf{SSIM} $\uparrow$ & \textbf{LPIPS} $\downarrow$ \\
        \midrule
        \multirow{4}{*}{\rotatebox{90}{\textbf{WildRGB-D}}} 
        & MotionCtrl & 12.394 & 0.293 & 0.404 \\
        & ZeroNVS  & 16.143  & 0.460  & 0.283 \\
        & ViewCrafter  & 13.960  & 0.375  & 0.290 \\
        & \textbf{LRAS (Ours)} & \textbf{17.748}  & \textbf{0.536} & \textbf{0.218} \\
        \midrule
        \multirow{4}{*}{\rotatebox{90}{\textbf{DL3DV}}} 
        & MotionCtrl & 12.629 & 0.261 & 0.462 \\
        & ZeroNVS  & 15.622  & 0.403  & 0.331 \\
        & ViewCrafter  & 16.592  & 0.430  & \textbf{0.253} \\
        & \textbf{LRAS (Ours)} & \textbf{18.110} & \textbf{0.523} & 0.328 \\
        \bottomrule
    \end{tabular}
    \caption{\textbf{Comparison of metrics for novel view synthesis.}}
    \label{tab:nvs}
    \vspace{-4.7mm}
\end{table}

 % on WildRGB-D and DL3DV datasets.

% \begin{table}[h]
%     \centering
%     \renewcommand{\arraystretch}{1.2} % Adjust row height for readability
%     \setlength{\tabcolsep}{6pt} % Adjust column spacing
%     \begin{tabular}{lcccc}
%         \toprule
%         \multirow{2}{*}{\textbf{Model}} & \multicolumn{2}{c}{\textbf{WildRGB-D}} & \multicolumn{2}{c}{\textbf{DL3DV}} \\
%         \cmidrule(lr){2-3} \cmidrule(lr){4-5}
%          & \textbf{PSNR} $\uparrow$ & \textbf{SSIM} $\uparrow$ & \textbf{PSNR} $\uparrow$ & \textbf{SSIM} $\uparrow$ \\
%         \midrule
%         ZeroNVS      & 21.12  & 0.874  & 22.35  & 0.891 \\
%         ViewCrafter  & 22.98  & 0.895  & 23.47  & 0.902 \\
%         \midrule
%         \textbf{PSI (Ours)} & \textbf{24.12} & \textbf{0.910} & \textbf{25.01} & \textbf{0.920} \\
%         \bottomrule
%     \end{tabular}
%     \caption{Comparison of Novel View Synthesis models on WildRGB-D and DL3DV datasets. Higher is better for PSNR and SSIM ($\uparrow$).}
%     \label{tab:nvs}
% \end{table}

\textbf{Evaluation Details}
To ensure a fair evaluation of novel view synthesis (NVS) on out-of-distribution datasets, we selected two benchmarks: WildRGB-D for object-centric NVS and DL3DV for scene-level NVS. For WildRGB-D, we randomly sampled 100 scenes from its evaluation split. For DL3DV, since some models were trained on this dataset, we selected 100 scenes from its recently released 11K subset, which, to the best of our knowledge, was not used to train any of the compared models. From each video, we extracted a 25-frame sequence and used the first frame as the input image and evaluating the generated frames. For quantitative evaluation, we measured PSNR, SSIM, and LPIPS \cite{zhang2018unreasonable}. As baselines, we compared against MotionCtrl, ZeroNVS, and ViewCrafter. Further implementation details are provided in the supplementary materials.

% Our model achieves high-quality novel view reconstruction, preserving object and scene identity with great controllability. 
% Figure \ref{fig:nvs} presents qualitative examples on the DL3DV and WildRGB-D datasets, highlighting our model’s reconstruction quality and camera controllability.

\textbf{Qualitative and Quantitative Comparisons} 
As shown in Figure \ref{fig:nvs}, our model achieves high-quality novel view reconstruction that maintains object and scene identity. In contrast, MotionCtrl distorts the scene and objects inconsistently. ZeroNVS often suffers from inaccurate 3D reconstruction and artifacts, and its hallucinated regions often become blurry and unrealistic. ViewCrafter may produce visually appealing images, but it frequently changes object appearance and global illumination. Our approach, built on local token-based autoregressive transformers, ensures object and global scene identity remain consistent. Our model also demonstrates robust and precise camera control, a significant advantage over previous methods. Despite our efforts to optimize scene scales, MotionCtrl fails to accurately control camera motion regardless of conditioning, while ZeroNVS faces pose alignment issues when its 3D reconstruction quality is poor. In contrast, our model directly computes pixel correspondences from depth, providing more intuitive and reliable control over camera motion. The key difference in scene alignment between ViewCrafter and our method is that the former optimizes scale in 3D point cloud space, whereas our method performs scale optimization directly in pixel space using optical flow correspondences.

% Despite our efforts to optimize scene scales, MotionCtrl fails to accurately control camera motion regardless of conditioning, while ZeroNVS faces pose alignment issues when its 3D reconstruction quality is poor. In contrast, our model directly operates on pixel correspondences, providing a more intuitive and reliable control over camera motion. The key difference from ViewCrafter is that it optimizes scale in the 3D point cloud space through scene alignment, whereas our method performs scale optimization directly in the pixel space using optical flow or correspondences.

% 1. erase that sentence
% 2. in contrast first, then explain viewcrafter is kind of equivalent
% 3. argue our method is better

Quantitatively, our model outperforms previous methods in reconstruction quality metrics, as shown in Table \ref{tab:nvs}. It achieves the best overall metrics on WildRGB-D, and the best PSNR and SSIM scores on DL3DV, although ViewCrafter attains a better LPIPS. These results reflect our model's performance with precise camera control and preservation of scene and object identity.

% This is because slightly blurred reconstructions under large physical camera motions.

% Overall, our model surpasses previous methods in NVS quality, effectively addressing common issues observed in diffusion-finetuned models.

\subsection{3D Object Manipulation}\label{subsec:obj_motion}

% \twocolumn[{
%     \maketitle
%     \begin{center}
%         \centering
%         \hspace*{-0.5cm}\includegraphics[width=\textwidth]{figures/object_motion.png}
%         \caption{We introduce physical scene editing.}
%         \label{fig:object_motion}
%     \end{center}
%     \vspace{0.3cm}
% }]

\begin{figure*}[t]
    \centering
    \includegraphics[width=.97\textwidth]{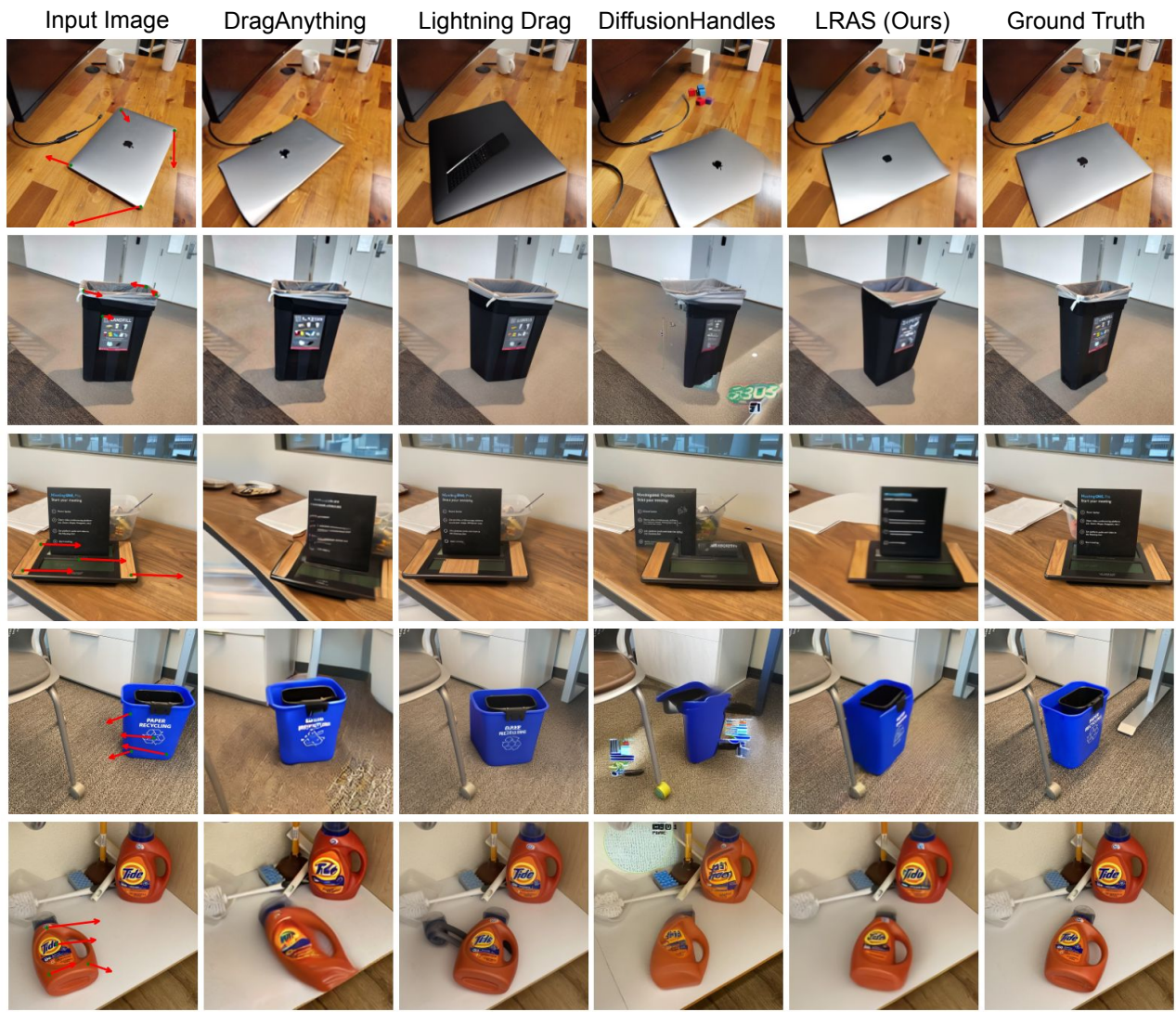}
    \captionsetup{labelfont=bf}
    \caption{\textbf{3D object manipulation from a single image.} We show that our model can perform both 3D object translation and rotation. Compared to the other methods, our model preserves object identity on real world images, and produces more photorealisic generated images with accurate object edits. See supplementary for more results. }
    \label{fig:object_motion}
    \vspace{-0.4cm}

\end{figure*}
% \vspace{-10pt}

\textbf{Baselines} We compare to DiffusionHandles~\cite{pandey2024diffusion}, which is the closest related work that performs 3D object edits using depth-conditioned diffusion models. Additionally, we also compare to drag-based image editing models such as LightningDrag~\cite{shi2024lightningdrag} and DragAnything~\cite{draganything}. Although these methods cannot be directly conditioned on 3D transforms, we find that providing sparse 2D flow vectors (which are part of our dataset's annotations) can be used to make these models work reasonably well for 3D manipulations. 

\textbf{New Object Editing Benchmark}
Most prior work in this area either use human evaluations on a small set of images~\cite{michel2023object}, or synthetic benchmarks~\cite{pandey2024diffusion} to evaluate their method. This can be attributed to the lack of high quality real-world datasets with ground-truth 3D object transform annotations. To address this problem, we collect a dataset called \textbf{\texttt{3DEditBench}} consisting of 100 image pairs with a diverse set of object types undergoing rotations and translations, and inter-object occlusions. We capture these images in a variety of background and lighting conditions. To obtain the ground-truth 3D object transformation for a given pair, we annotate four corresponding points in the two images, unproject them, and use least-squares optimization to find the best-fitting rigid transformation that aligns the two sets of points. This transform is then used to create flow maps that condition \lrasrgb to perform 3D object edits in natural scenes (see Section.~\ref{subsec:obj_motion})

\textbf{Metrics} In line with our NVS evaluations in Section.~\ref{subsec:nvs}, we use metrics that measure the image generation quality such as PSNR, SSIM and LPIPS. However, previous work~\cite{pandey2024diffusion} has found that these metrics often prefer image quality over edit accuracy.~\cite{pandey2024diffusion} proposed the Edit Adherance metric (EA) to directly measure of how well the boundaries of the transformed object overlaps with the ground truth. This is measured as the IOU (intersection over union) between the ground truth segment map and the estimated segment map in the generated image -- we obtain these by running the SAM~\cite{kirillov2023segment} model on these images.

\begin{table}[b]
    \centering
    \hspace*{-0.4cm} % Adjust the value as needed
    \renewcommand{\arraystretch}{1.2} % Adjust row height for readability
    \setlength{\tabcolsep}{5pt} % Adjust column spacing
    \begin{tabular}{lccccc}
        \toprule
        \textbf{Model} &  \textbf{PSNR}  $\uparrow$ &   \textbf{SSIM} $\uparrow$ & \textbf{LPIPS} $\downarrow$ & \textbf{EA} $\uparrow$ \\
        \midrule
        DragAnything  &	15.13 & 0.415 & 0.443 & 0.517 \\
        Diffusion Handles & 17.82 & 0.567 &	0.344 & 0.619  \\
        LightningDrag & 19.52 & 0.567 & \textbf{0.184} & 0.722  \\        
        \textbf{LRAS (Ours)} & \textbf{21.85 } & \textbf{ 0.700} & 0.212 & \textbf{0.798}  \\
        \bottomrule
        
    \end{tabular}
    % \vspace{3mm}
    \captionsetup{labelfont=bf}
    \caption{\textbf{Comparison of metrics for 3D object manipulation.}}
    \label{tab:object_motion}
    \vspace{-0.4cm}
\end{table}

% \begin{table}[h]
%     \centering
%     \renewcommand{\arraystretch}{1.2} % Adjust row height for readability
%     \setlength{\tabcolsep}{6pt} % Adjust column spacing
%     \begin{tabular}{lcccc}
%         \toprule
%         \textbf{Model} &  MSE $\downarrow$ &  PSNR  $\uparrow$ & LPIPS $\downarrow$  &  SSIM $\uparrow$  \\
%         \midrule
%         DragAnything  & 0.045 &	14.31 &  0.405	& 0.469 \\
%         Diffusion Handles & 0.023 & 16.75 &	0.339 & 0.613	  \\
%         LightningDrag & 0.017 & 18.97 & \textbf{0.167 }& 0.741  \\        
%         \textbf{LRAS (Ours)} & \textbf{0.009} & \textbf{21.08 } & 0.193 & \textbf{0.770 }  \\
%         \bottomrule
        
%     \end{tabular}
%     \vspace{3mm}
%     \captionsetup{labelfont=bf}
%     \caption{\textbf{Comparison of metrics for 3D object manipulation.}}
%     \label{tab:object_motion}
% \end{table}

\textbf{Qualitative and Quantitative Comparisons}
We find that our model outperforms other methods on all metrics except marginally inferior LPIPS compared to LightningDrag (see Table.~\ref{tab:object_motion}). However, as shown in Figure.~\ref{fig:object_motion}, we find that qualitatively our model is significantly better, especially on more complex 3D transformations. Furthermore, the Edit Adherence (EA) metric (proposed in ~\cite{pandey2024diffusion}), which is a more reliable measure of the precision of the edit, seems to strongly prefer generations of our model. Interestingly, we find that DiffusionHandles~\cite{pandey2024diffusion} struggles on some of these real-world images due to failures in the null-text inversion process for natural images. The failure modes involve changing the appearance of the surrounding objects in the scene, leading to unnatural generations, blurry reconstructions, and incorrect 3D motion. A similar trend can also be seen in the drag-based image-editing baselines, albeit to a lesser degree in LightningDrag. On the other hand, \lras overcomes these limitations with autoregressive sequence modeling and generates more consistent and natural-looking images. Further, we find that our method can also be extended to perform object removal and amodal completion (we show more examples in supplementary).

% Surprisingly, LightningDrag outperforms diffusion handles on this task even though it gets inferior inputs in the form of sparse 2D vectors. 

% \subsection{Random access flow model}
\subsection{End-to-End 3D Scene Understanding}

\subsubsection{Self-Supervised Monocular Depth Estimation}

% \twocolumn[{
%     \maketitle
%     \begin{center}
%         \centering
%         \hspace*{-0.5cm}\includegraphics[width=\textwidth]{figures/depth.png}
%         \caption{The}
%         \label{fig:depth}
%     \end{center}
%     \vspace{0.3cm}
% }]

\begin{figure}[t]
    \centering
    \includegraphics[width=0.48\textwidth]{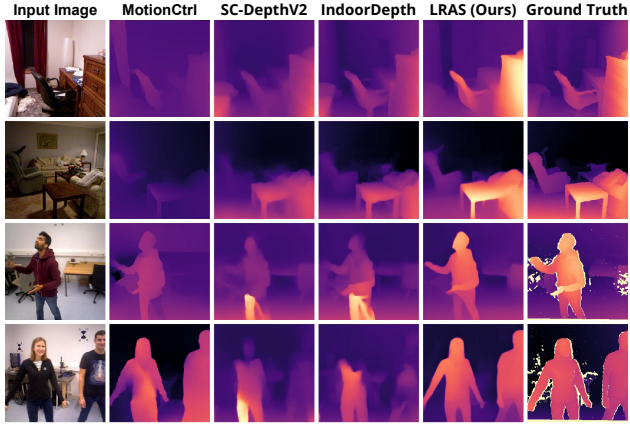}
    \captionsetup{labelfont=bf}
    \caption{\textbf{Self-supervised monocular depth estimation.} On static scenes, our model performs comparably well to existing self-supervised depth estimation methods. However, when there are dynamic objects in the scene, our model significantly outperforms geometric-consistency-based methods, demonstrating its robustness in handling moving objects. Yellow artifacts in ground truth depth maps are noise and excluded during the evaluation.}
    \label{fig:depth}
\end{figure}

\begin{table}[t]
    \centering
    \hspace*{-0.4cm} % Adjust the value as needed
    \renewcommand{\arraystretch}{1.0} % Adjust row height for readability
    \setlength{\tabcolsep}{6pt} % Adjust column spacing
    \begin{tabular}{clccc}
        \toprule
        \textbf{Dataset} & \textbf{Model} & \textbf{AbsRel} $\downarrow$ & \textbf{Log10} $\downarrow$ & \textbf{$\delta_1$} $\uparrow$ \\
        \midrule
        \multirow{4}{*}{\rotatebox{90}{\textbf{NYUD-v2}}} 
        & MotionCtrl   & 0.246  & 0.099 & 0.624 \\
        & SC-DepthV2  & 0.136  &  0.059  &  0.819  \\
        & IndoorDepth  & \textbf{0.120}  &   0.051 & 0.857 \\
        & \textbf{LRAS (Ours)} & 0.121  &    \textbf{0.050} & \textbf{0.873} \\
        \midrule
        \multirow{4}{*}{\rotatebox{90}{\textbf{BONN}}} 
        & MotionCtrl   & 0.167  & 0.068 & 0.798   \\
        & SC-DepthV2  & 0.183  &  0.169  &  0.800   \\
        & IndoorDepth  & 0.167  &  0.064  & 0.827  \\
        & \textbf{LRAS (Ours)} & \textbf{0.120}  &  \textbf{0.047}  &  \textbf{0.889} \\
        \midrule
        \multirow{4}{*}{\rotatebox{90}{\textbf{TUM}}} 
        & MotionCtrl   & 0.204  & 0.097  & 0.712  \\
        & SC-DepthV2  & 0.229  &   0.100  &  0.632  \\
        & IndoorDepth  & 0.213  &   0.094 &   0.682  \\
        & \textbf{LRAS (Ours)} & \textbf{0.179}  &   \textbf{0.073}  &   \textbf{0.766}  \\
        \bottomrule
    \end{tabular}
    \captionsetup{labelfont=bf}
    % \hspace*{-0.3cm}
    \caption{\textbf{Comparison of metrics for self-supervised monocular depth estimation.}}
    \label{tab:depth_comparison}
    \vspace{-0.4cm}

\end{table}

\textbf{Evaluation Details}
We evaluate the self-supervised monocular depth estimation performance on three datasets: NYUv2 \cite{silberman2012indoor}, BONN \cite{palazzolo2019refusion}, TUM \cite{sturm2012benchmark} datasets. NYUv2 is mostly composed of static scenes, whereas BONN and TUM include humans with implied motion. We evaluate SC-DepthV2 \cite{bian2021auto}, IndoorDepth \cite{fan2023deeper}, and MotionCtrl \cite{wang2024motionctrl} as baselines. To extract depth from MotionCtrl, we induce an upward in-plane camera motion and compute the disparity between the first and 7th images using RAFT ~\cite{teed2020raft}. As our model uses square images as input, we center-crop the images for all datasets and only evaluate the square region.

\textbf{Qualitative and Quantitative Comparisons}
Our results demonstrate that \lras achieves high-quality depth reconstruction in both static and dynamic settings, as shown in Figure \ref{fig:depth}. The baseline models exhibit limitations because they rely on static geometry consistency, preventing them from extracting training signals from moving objects. 
In contrast, our model successfully learns depth cues from optical flow. 
While optical flow is not always purely induced by camera motion, we empirically found that averaging optical flow while moving the camera upward leads to reliable depth estimation. MotionCtrl demonstrates better generalization to dynamic objects than other self-supervised methods, but lacks strong depth understanding overall, as evident by its weaker performance in static scenes. Table \ref{tab:depth_comparison} confirms our observations quantitatively, where our model achieves competitive performance on NYUv2, and outperforms other methods on dynamic datasets, BONN and TUM. Overall, our findings in self-supervised depth estimation highlight the importance of optical flow as a prediction target. The results also strengthen the argument for autoregressive modeling, where simple modification of sequence design can naturally facilitate other tasks.

% Quantitatively, Table \ref{tab:depth_comparison} confirms the qualitative observations. Our model achieves competitive performance on NYUv2, demonstrating that it captures depth information effectively without explicitly enforcing geometric consistency. This supports the argument for leveraging more flexible representations, such as optical flow, instead of strictly adhering to geometric consistency constraints. Moreover, while other SC-DepthV2 and IndoorDepth models were trained on NYUv2, our model achieves these results in a zero-shot setting. On dynamic data, it significantly outperforms alternatives, indicating a superior ability to handle depth estimation for moving objects, as reflected in qualitative examples.

\subsubsection{End-to-end Novel View Synthesis}
\vspace{-0.4cm}

% \begin{table}[h]
%     \centering
%     \renewcommand{\arraystretch}{1.2} % Adjust row height for readability
%     \setlength{\tabcolsep}{4pt} % Adjust column spacing
%     \begin{tabular}{lccc}
%         \toprule
%         \textbf{Dataset} \\
%         \textbf{Model} & \textbf{PSNR} $\uparrow$ & \textbf{SSIM} $\uparrow$ & \textbf{LPIPS} $\downarrow$ \\
%         \midrule
%         \textbf{WildRGB-D} \\
%         LRAS \newline w. Our Depth  & 16.903 & 0.491 & 0.257 \\
%         LRAS \newline w. DepthAnythingv2 & 17.895  & 0.541 & 0.231 \\
%         \midrule
%         \textbf{DL3DV} \\
%         Ours w. Our Depth  & 18.169  & 0.524  & 0.324 \\
%         Ours w. DepthAnythingv2 & 18.279 & 0.531 & 0.321 \\
%         \bottomrule
        
%     \end{tabular}
%     \captionsetup{labelfont=bf}
%     \caption{\textbf{Comparison of metrics for novel view synthesis depending on depth model.}}
%     \label{tab:nvs_our_depth}
% \end{table}

\begin{table}[h]
    \centering
    \hspace*{-0.4cm}
    \renewcommand{\arraystretch}{1.2} % Adjust row height for readability
    \setlength{\tabcolsep}{4pt} % Adjust column spacing
    \begin{tabular}{llccc}
        \toprule
        \textbf{Dataset} & \textbf{LRAS} & \textbf{PSNR} $\uparrow$ & \textbf{SSIM} $\uparrow$ & \textbf{LPIPS} $\downarrow$ \\
        \midrule
        \multirow{2}{*}{\textbf{WildRGB-D}} 
        & \makecell[l]{w. Our Depth}  & 16.716 & 0.484 & 0.264 \\
        & \makecell[l]{w. DA-V2} & \textbf{17.748}  & \textbf{0.536} & \textbf{0.218} \\
        \midrule
        \multirow{2}{*}{\textbf{DL3DV}} 
        & \makecell[l]{w. Our Depth}  & 17.984  & 0.516  & 0.332 \\
        & \makecell[l]{w. DA-V2} & \textbf{18.110} & \textbf{0.523} & \textbf{0.328} \\
        \bottomrule
    \end{tabular}
    \captionsetup{labelfont=bf}
    \caption{\textbf{Comparison of metrics for novel view synthesis depending on depth model.}}
    \label{tab:nvs_our_depth}
    \vspace{-0.2cm}
\end{table}

% \begin{table}[h]
%     \centering
%     \renewcommand{\arraystretch}{1.2} % Adjust row height for readability
%     \setlength{\tabcolsep}{6pt} % Adjust column spacing
%     \begin{tabular}{lcccc}
%         \toprule
%         \multirow{2}{*}{\textbf{Model}} & \multicolumn{2}{c}{\textbf{WildRGB-D}} & \multicolumn{2}{c}{\textbf{DL3DV}} \\
%         \cmidrule(lr){2-3} \cmidrule(lr){4-5}
%          & \textbf{PSNR} $\uparrow$ & \textbf{SSIM} $\uparrow$ & \textbf{PSNR} $\uparrow$ & \textbf{SSIM} $\uparrow$ \\
%         \midrule
%         ZeroNVS      & 21.12  & 0.874  & 22.35  & 0.891 \\
%         ViewCrafter  & 22.98  & 0.895  & 23.47  & 0.902 \\
%         \midrule
%         \textbf{PSI (Ours)} & \textbf{24.12} & \textbf{0.910} & \textbf{25.01} & \textbf{0.920} \\
%         \bottomrule
%     \end{tabular}
%     \caption{Comparison of Novel View Synthesis models on WildRGB-D and DL3DV datasets. Higher is better for PSNR and SSIM ($\uparrow$).}
%     \label{tab:nvs}
% \end{table}

Since our framework can estimate depth, we explored using our model instead of a supervised depth model to create a fully self-supervised NVS pipeline. Table \ref{tab:nvs_our_depth} presents the results of NVS using depth predicted by our model. While the metrics generally declined compared to the pipeline using Depth Anything V2 (DA-V2), the drop was not severe. This indicates that our model's depth estimation is sufficiently accurate for novel view synthesis, reinforcing the feasibility of a unified, self-supervised 3D vision framework with optical flow and autoregressive training.

% \section{Ablations}

% \input{sections/ablations}

% \section{Discussion}

% \input{sections/discussion}

\section{Discussion \& Conclusion}

In this work, we introduce \lras, an autoregressive sequence modeling framework with local patch quantization and random access prediction. We show that our method outperforms diffusion-based models in 3D editing capabilities, ensuring consistency in objects and scenes during editing. The model also offers precise camera control and object manipulation, demonstrating a strong understanding of spatial relationships and transformations in 3D. Furthermore, we demonstrate that our modeling framework is flexible. With a simple change in sequence design, it can leverage optical flow either as input conditioning for 3D scene editing or as a prediction target for depth estimation. 
% Additionally, our results in self-supervised depth estimation further establish optical flow as a powerful and fundamental representation for 3D vision tasks.

Overall, \lras provides a robust and scalable alternative to diffusion models for 3D scene understanding, expanding the potential of autoregressive modeling in vision. Future work could explore the integration of additional modalities to further enhance spatial and physical reasoning.

\section{Acknowledgment}

This work was supported by the following awards: To D.L.K.Y.: Simons Foundation grant 543061, National Science Foundation CAREER grant 1844724, National Science Foundation Grant NCS-FR 2123963, Office of Naval Research grant S5122, ONR MURI 00010802, ONR MURI S5847, and ONR MURI 1141386 - 493027. We also thank the Stanford HAI, Stanford Data Sciences and the Marlowe team, and the Google TPU Research Cloud team for computing support.

%-------------------------------------------------------------------------

{
    \small
    \bibliographystyle{ieeenat_fullname}
    \bibliography{main}
}

% \input{sec/X_suppl}
% \clearpage
% \setcounter{page}{1}
% \maketitlesupplementary

\clearpage
\maketitlesupplementary
\appendix
\renewcommand{\thesection}{\Alph{section}}
\setcounter{section}{0} % Resets section counter for alphabetical ordering

\section{Dataset} \label{subsec:sup_data}
We collect a large dataset of diverse video clips crawled from the internet, totaling about 7,000 hours in length, called big video dataset.

The videos were crawled using LLaMA 3 \cite{Dubey2024TheL3}-generated search queries about videos that contain lots of physical dynamics, diverse settings and objects. Specifically, the crawl search queries were generated using Kinetics400 \cite{kay2017kinetics} action categories, and supplemented with additional sport and physical activity categories, as well as product review categories. The videos were filtered to contain some minimal amount of optical flow, and to align with predefined CLIP \cite{radford2021clip} keyword filters. Our positive CLIP keywords are
``action,'' ``activity,'' ``motion,''  and ``place,'' and our negative keywords are ``animation,'' ``cartoon,'' ``face,'' ``game menu,'' ``graphic,'' ``map,'' ``newscast,'' ``person,''  and ``screenshot.'' Alignment is measured as the dot product between the CLIP embeddings of keywords and video frames.

To improve camera motion diversity in the \lrasflow training data, we converted 280 scenes from ScanNet++ \cite{yeshwanth2023scannet++} into Neural Radiance Fields \cite{mildenhall2021nerf,nerfstudio} and rendered videos from them with known diverse camera trajectories. Discretized relative camera pose change between two frames is provided to \lrasflow as conditioning when available in the data.

\section{NVS Evaluation Details}
\begin{figure}[t]
    \centering
    \includegraphics[width=\columnwidth]{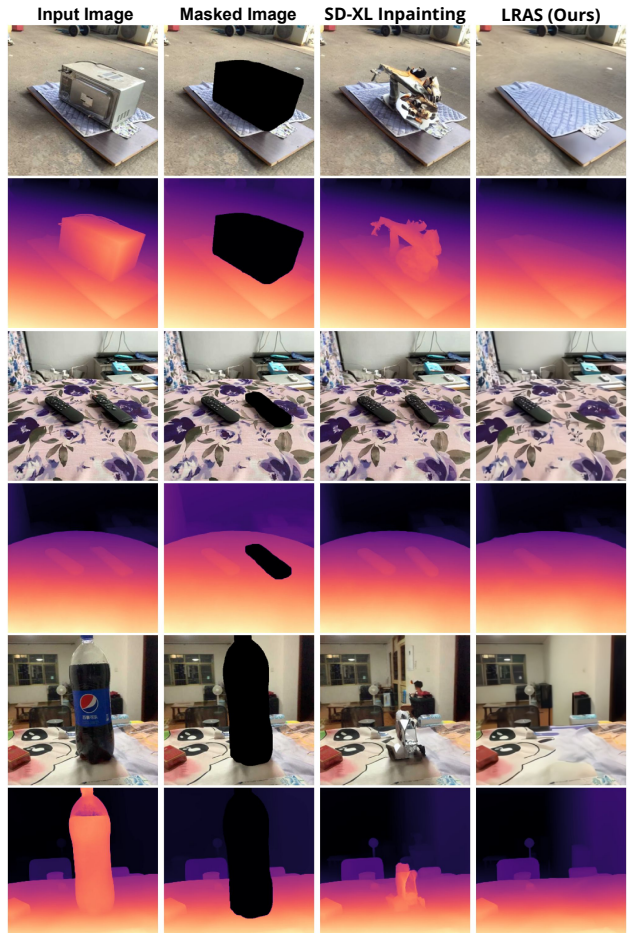}
    \captionsetup{labelfont=bf}
    \caption{\textbf{Amodal Completion.} We compare our self-supervised amodal depth reasoning with the inpainting method. The inpainting method struggles with underdetermined scene changes, as it lacks explicit control over object removal. In contrast, the flow-based physical scene editing approach conditions object removal more precisely, resulting in more reliable amodal reasoning.}
    \label{fig:amodal_depth}
\end{figure}
To evaluate novel view synthesis, we compare generated images to ground-truth real-world images using known camera poses. While camera rotation is unambiguous, camera translation may have arbitrary scale. Therefore, it is necessary to find the right scene scale to perform fair evaluations for all of the models.

To align MotionCtrl and ZeroNVS results with ground-truth images, we sweep a range of scene scales and take the generated trajectories with the best median LPIPS score across frames. For ZeroNVS, we sweep scales in the range 0.1 to 10, multiplying the scale by the ground-truth camera translations from each evaluation dataset. ZeroNVS introduces a normalization scheme \cite{sargent2023zeronvs} at training time to address this scale ambiguity, but does not apply it at inference. For MotionCtrl, we sweep the range 1 to 10, as smaller translation scales empirically weaken the camera conditioning and lead to incorrect camera pose trajectories. Scale alignment for these models may fail for samples with especially poor 3D reconstruction quality. For ViewCrafter, we resolve the scene scale using their method of aligning point clouds with DUSt3R \cite{wang2024dust3r}. For \lras, we have computed the single scale value per scene by matching the optical flow computed from the video using RAFT \cite{teed2020raft} and the 2D flow computed from the depth and relative camera pose changes.

Since ViewCrafter operates on wide rectangular videos, we adapt the input images accordingly. For DL3DV, which consists of wide images, we provide the full image to ViewCrafter. For WildRGB-D, which contains narrower images, we provide a center-cropped rectangular region to ViewCrafter. All other models receive a center-cropped square image as input for both datasets. All evaluation metrics are computed only on the overlapping regions; for WildRGB-D, this region is rectangular, and for DL3DV it is square.

\section{Amodal Completion}\label{subsec:amodal}

A simple yet powerful application of our model is amodal reasoning. By applying high-magnitude flow to the object, we effectively remove it from the scene. We compare this approach to a self-supervised heuristic for object removal based on image inpainting using the Stable Diffusion XL (SD-XL) \cite{podell2023sdxl} model. As shown in Fig. \ref{fig:amodal_depth}, our model successfully removes objects while reconstructing the occluded regions with reasonable accuracy. In contrast, the SD-XL approach may struggle with imperfect segmentation or implicit object presence caused by shadows or nearby objects. This problem is also observed by other work \cite{winter2024objectdrop}, where they address it by introducing a specific dataset for training. Our method, however, provides an explicit physical cue for object removal via optical flow, enabling more controlled and interpretable amodal reasoning in a self-supervised way.

\section{Additional Qualitative results on NVS and object manipulations}

\begin{figure*}
    \centering
    \includegraphics[width=0.97\textwidth]{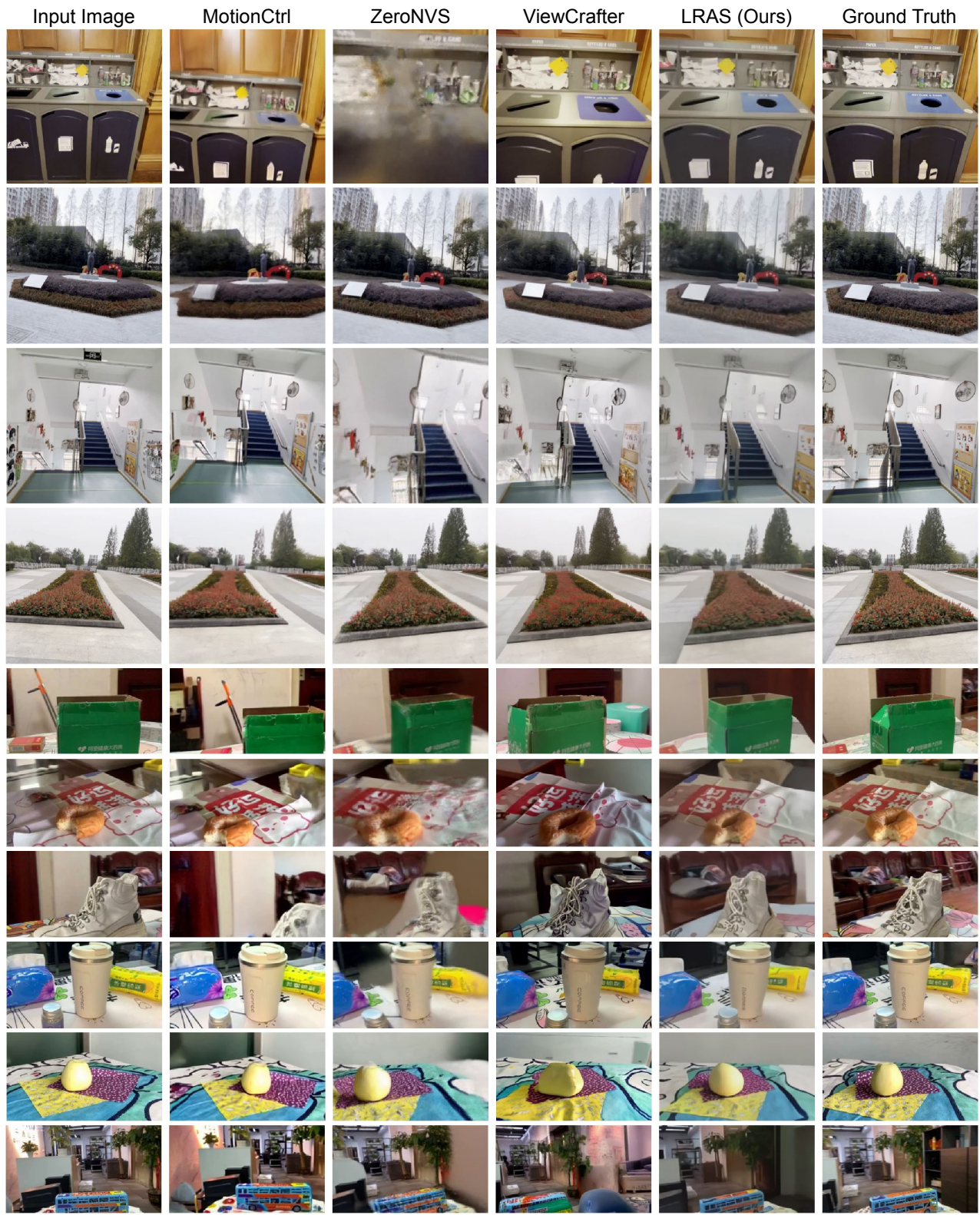}
    \captionsetup{labelfont=bf}
    \caption{\textbf{Additional results on novel view synthesis from a single image.} The results show that our model performs controllable novel view synthesis with various camera motions in a diverse scenes. Compared to other models, the reconstructed images do not show abrupt change in object and scene identity.}
    \label{fig:nvs_sup}
    \vspace{-0.4cm}

\end{figure*}

In Figure~\ref{fig:nvs_sup} we include additional qualitative results for novel view synthesis and in Figure~\ref{fig:object_motion} we include additional qualitative results for object manipulation.

\begin{figure*}[t]
    \centering
    \includegraphics[width=\textwidth]{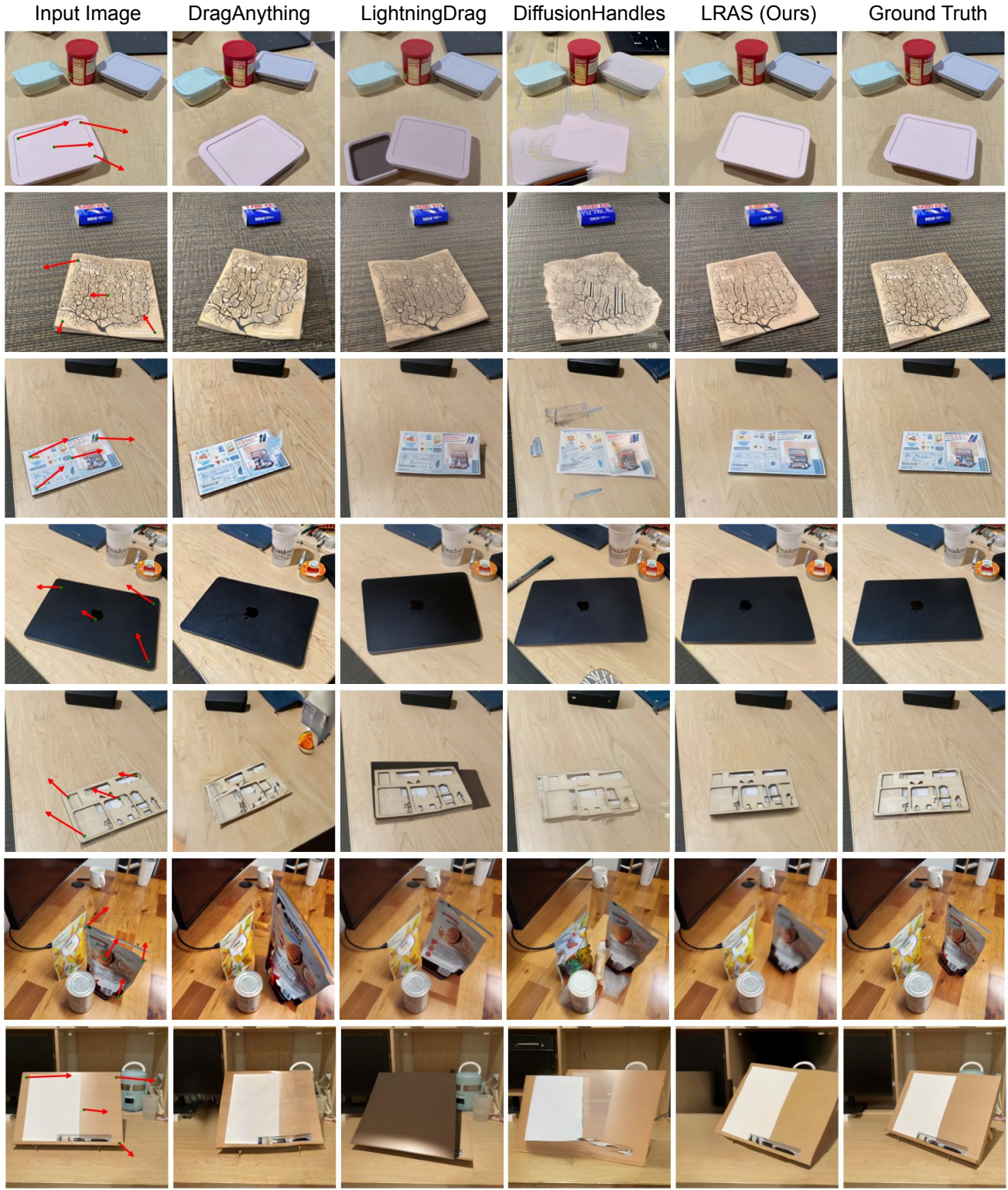}
    \captionsetup{labelfont=bf}
    \caption{\textbf{Additional results on 3D object manipulation from a single image.} The results show that our model can perform both object translation and rotation in 3D. Compared to the other methods, our model does not change the object identity even for in-the-wild real world images.}
    \label{fig:object_motion}
    \vspace{-0.4cm}

\end{figure*}

%-------------------------------------------------------------------------

% {\small
% \bibliographystyle{ieeenat_fullname}
% \bibliography{main}
% }

\end{document}